\documentclass[10pt,conference,a4paper]{IEEEtran}

\usepackage{hyperref}
\usepackage{multirow}
\usepackage{siunitx}
\usepackage{graphicx}
\usepackage{subcaption}

\usepackage{amsmath}
\usepackage{amssymb}
\usepackage[ruled,vlined]{algorithm2e}
\usepackage{sidecap}
\usepackage[export]{adjustbox}
\usepackage{makecell}
\usepackage[dvipsnames, table]{xcolor}

\begin{document}
	%
	\title{Text Recognition - \\
	Real World Data and Where to Find Them}


		\author{\IEEEauthorblockN{Klara Janouskova, Jiri Matas}
		\IEEEauthorblockA{Centre for Machine Perception, Department of Cybernetics \\
			Czech Technical University, Prague, Czech Republic}
		\and

		\IEEEauthorblockN{Lluis Gomez, Dimosthenis Karatzas}
		\IEEEauthorblockA{Computer Vision Center\\
			Universitat Autonoma de Barcelona, Spain\\}
		
	}

	\maketitle
	
	
	\begin{abstract}
		We present a method for exploiting weakly annotated images to improve text extraction pipelines. The approach uses an arbitrary end-to-end text recognition system to obtain text region proposals and their, possibly erroneous, transcriptions. The proposed method includes matching of imprecise transcription to weak annotations and edit distance guided neighbourhood search. It produces nearly error-free, localised instances of scene text, which we treat as ``pseudo ground truth'' (PGT).
		
		We apply the method to two weakly-annotated datasets. Training with the extracted PGT consistently improves the accuracy of a state of the art recognition model, by 3.7~\% on average, across different benchmark datasets (image domains) and 24.5~\% on one of the weakly annotated datasets.

		
		
	\end{abstract}


	%
	\IEEEpeerreviewmaketitle

	\section{Introduction}
	
	Written text is an important source of information for humans and plays an important role in everyday life, being frequently present in scenes with man-made structures. It is one of the most common classes in general object detection datasets like CoCo~\cite{veit2016coco}. Research in end-to-end reading
	systems is a very active research field 
	in academia and industry alike.
	It is an essential part of many applications ranging from translation systems and autonomous driving to image retrieval or visual question answering.
	
	In recent years, with the introduction of deep neural network models, the field has advanced substantially. At the same time, state of the art performance comes at the cost of the requirement of large-scale annotated data for training.
	Such data need to be rich in geometry, style and content.
	Typical ground truth data is defined at the granularity of words as polygonal regions in the image along with the corresponding text transcriptions.
	The acquisition of such data requires substantial human effort and is very costly.
	
	The lack of data is usually approached in two different ways, either by generating synthetic data as in \cite{chen2019cross, gomez2019selective, zhan2019ga, gupta2016synthetic, liao2019synthtext3d, long2020unrealtext} or with different forms of weakly, semi or unsupervised learning on real data as in \cite{Baek_2019_CVPR, Sun_2019_ICCV, qin2019curved}.
	
		{
	\begin{figure}
		\centering
		\setlength{\tabcolsep}{3pt}
		\begin{tabular}{ccccc}
			\includegraphics[width=1.5cm, clip]{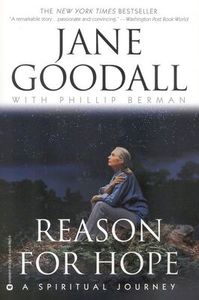} &
			\includegraphics[width=1.5cm, clip]{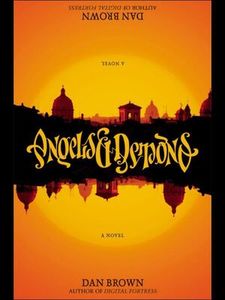} &
			\includegraphics[width=1.5cm, clip]{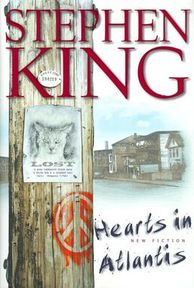} &
			\includegraphics[width=1.5cm, clip]{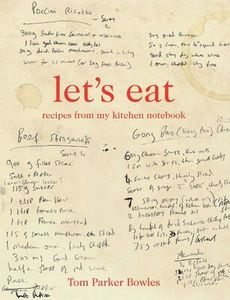} &
			\includegraphics[width=1.5cm, clip]{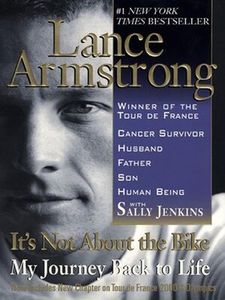} \\
			(a) & (b) & (c) & (d) & (e) \\
		\end{tabular}

\vspace{-1ex}
		\caption{The ABC Dataset images are diverse, some have 
		(a) both a simple layout and font,
		(b) a very artistic, almost illegible font and 
		(c) a font that resembles handwriting.
		Other have (d) both dense background and hand-written text  or (e) the background varies significantly even at the word level. The dataset includes weak annotations - the name of the author and the title. Not all text is annotated, location of the annotated text is not provided.}
		\label{fig:abc-examples}
		\vspace{-2ex}

	\end{figure}

}
    \begin{figure}

	\begin{subfigure}{0.95\columnwidth}
		\centering
		\includegraphics[width=\columnwidth]{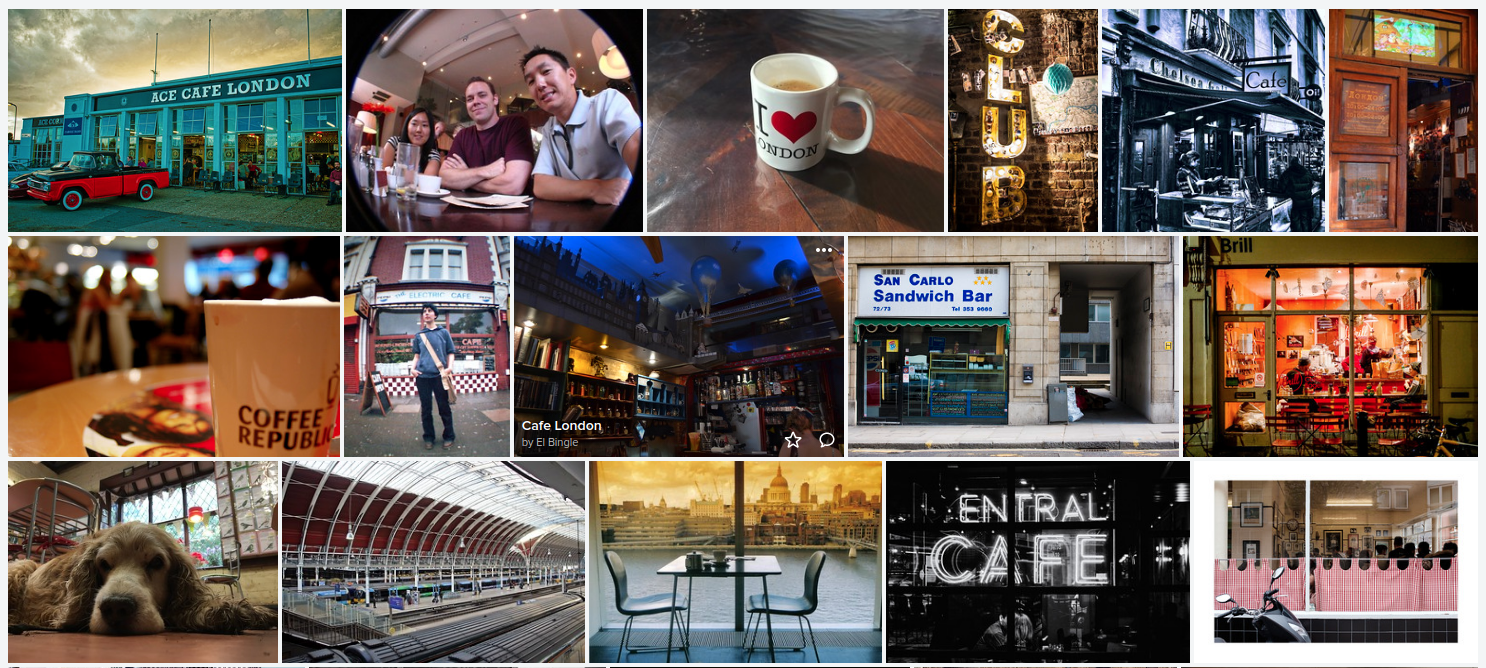}
					\vspace{-4ex}
		\subcaption{Searching for `Cafe London' photos
		in the Flickr application.}
							\vspace{1ex}
	\end{subfigure}  
	
	\begin{subfigure}{0.95\columnwidth}
			\setlength{\tabcolsep}{1.5pt}

		\centering
		\begin{tabular}{ccccc}
			\includegraphics[width=1.55cm]{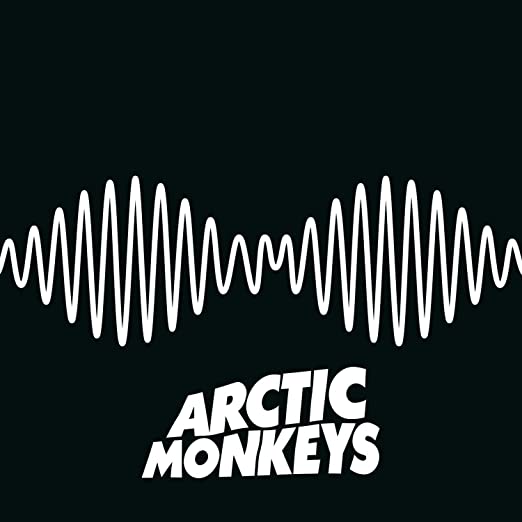} &
			\includegraphics[width=1.75cm]{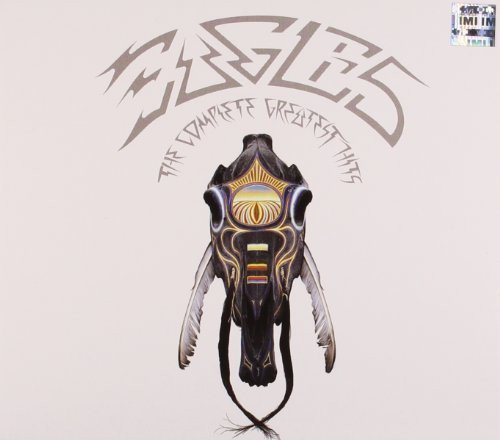} &
			\includegraphics[width=1.55cm]{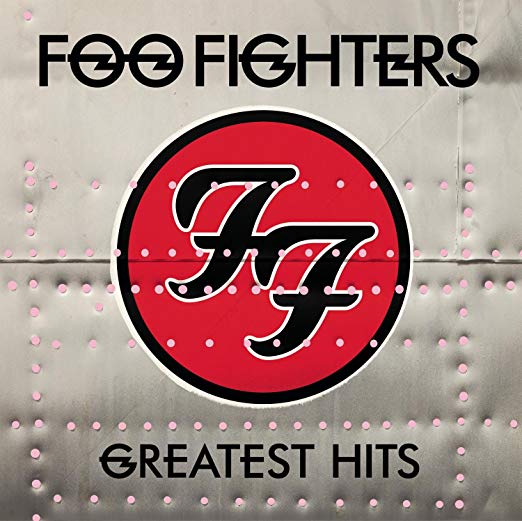} &
			\includegraphics[width=1.55cm]{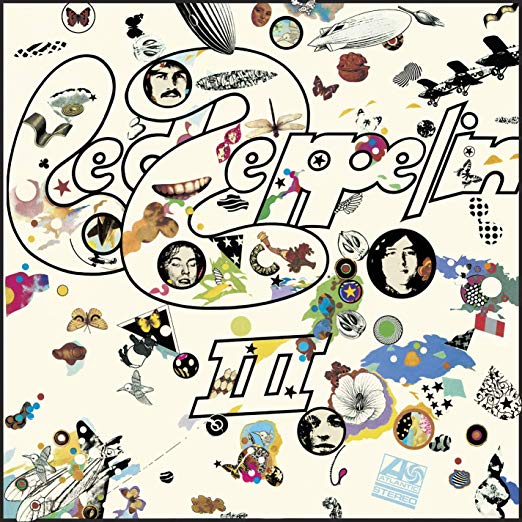} &
			\includegraphics[width=1.55cm]{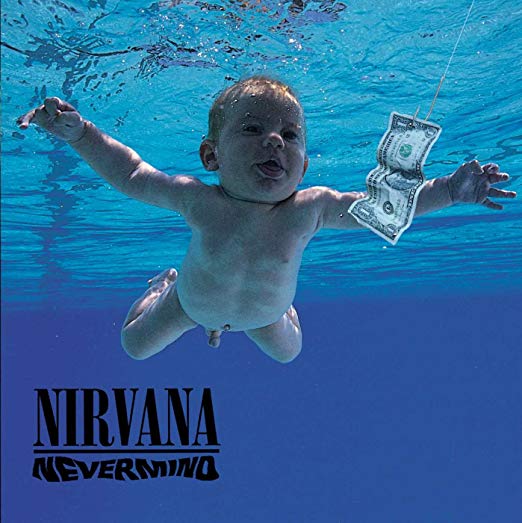} \\
		\end{tabular}
					\vspace{-2ex}
		\subcaption{Artist and album name in Amazon Music product database.}
	\end{subfigure}

	\caption{Potential sources of weakly annotated data; the availability of such data for research purposes changes over time.}
	\label{fig:weak_sources}
			\vspace{-2ex}

\end{figure}

	While fully annotated real world data are expensive and sparse, weakly annotated data in the form of images along with a set of words likely to appear in them are common.
	An example of a source for such weakly annotated data are product databases
	where we can readily obtain the name of the product and other meta-data. 
	A weakly annotated dataset, based on a product database, used in our experiments is shown in Figure \ref{fig:abc-examples}, other sources in Figure \ref{fig:weak_sources}.
	Images from mapping services like \href{https://www.google.com/maps}{Google Maps} are another potential source where street names and numbers or business names are words very likely to appear in an image and easily obtainable 
	through 
	location-based search. For example, given images from the location where a restaurant is supposed to be, it is 
	expected that the name of the restaurant will be visible in some of them

	In this paper, we present a new method for automatically generating pseudo ground truth (PGT) in the form of text regions and their transcription from weakly annotated data consisting of text transcriptions only with no information about the text location. The weak annotations may be noisy - incomplete (not all the text in the image is there) and with distractor words (text that is not in the image). The method requires an end-to-end text recognition system (E2E) pre-trained with 
	annotated data. 
	In contrary to our method which uses automatically obtained weak annotations, previous methods used weak labels which alleviate but do not eliminate human participation, such as annotating 
	the areas of interest.
	
	The core idea of our method is that the OCR output of the recognition model can be used to identify the most probable text match from the weak annotations by finding the one with the lowest edit distance. The detections that produce an exact match with the weak label are assumed to be correct. Furthermore, the recognition output can be used to find the modifications of the detected regions that minimize the edit distance to the matched text. For example, if we have predicted the word `car' and the best match was `cartoon', running the recognition again on the detected region extended to the right may decrease the distance between the matched and predicted text, possibly leading to the prediction of the matched word `cartoon'. We have experimentally verified that he probability of the recognition model giving the same output as the weak label for a wrong text region is very low, thus it is safe to use such regions as PGT for further training. 
	
	In summary, given an image and a dictionary of words as an input, the method outputs a subset of the dictionary words with their corresponding text regions in the image. The method is independent of the underlying implementation of the models.
	
	Possible applications of our method are improving the performance of an existing general E2E system or domain adaptation, where the source domain has full ground truth data
	whereas only weak labels are available for the target domain. 
	
	We apply our method to data from two different sources - a database of images of book-covers downloaded from Amazon books, using the title and the author of the book as weak annotations, and the Uber-Text dataset \cite{zhang2017uber}, which is very similar to the kind of data that could be obtained using a mapping system. 
	We train a recognition model with the PGT generated from both sources on various benchmarks, showing that it consistently improves the recognition accuracy across a wide range of datasets.
	We chose Uber-Text, a dataset with region-level annotations available, to evaluate how our approach compares to fully supervised training in ablation studies. Otherwise, the full annotations are not used. 
	
	The contributions of the paper are:
	\begin{itemize}
		\item We present a novel method
		for
		automatic generation of pseudo ground truth data
		from images with weak annotations.
		\item  The accuracy of the best-performing open-source model is improved significantly (3.7 \% on average) on a wide range of benchmark datasets by training with the PGT generated from two datasets.
		\item  The PGT 
		data significantly improve recognition performance
		in weakly-supervised domain adaptation (24.5 \%).
		\item  The PGT annotations in the Amazon Book Covers dataset will be published.
	\end{itemize}

	\section{Related Work}
	
	We first give a short introduction of scene text detection and recognition methods, and an overview of methods for generating synthetic data. Then we focus on weakly-supervised learning for text detection and recognition.
	
	\subsection{Text Detection and Recognition}
	Before the deep learning era, methods based on SWT or MSERs were used for text detection, for example~\cite{epshtein2010detecting, neumann2010method}.
	Subsequent 
	models were mostly based on region proposal approaches like \cite{jaderberg2016reading}. Recently, methods have rather turned to segmentation-based approaches like \cite{liao2019mask} and focused on representing arbitrarily shaped text, for example~\cite{Baek_2019_CVPR, long2018textsnake}. 
	
	Recent approaches for text recognition rely on deep learning. Most methods can be described by 4 stages.
	Transformation - a Spatial Transformer Network \cite{jaderberg2015spatial} normalizing the input image.
	Feature extraction - a CNN such as VGG~\cite{simonyan2014very} or ResNet \cite{he2016deep} maps the input image to feature maps.
	Sequence modelling - BiLSTMs are used to provide contextual information to the feature maps.
	Prediction - either CTC~\cite{Graves2006ConnectionistTC} or attention-based prediction \cite{cheng2017focusing} is used to convert the encoded features into a character sequence.
	Some methods treat the two tasks jointly, sharing features for both detection and recognition, for example~\cite{Liu_2018_CVPR, buvsta2018e2e, Qin_2019_ICCV}.

	\subsection{Synthetic data for text detection and recognition}
	
	The work of \cite{gupta2016synthetic} and \cite{jaderberg2016reading} had a great influence on the performance of text detection and recognition systems. 
%
	Synthetic data have proven to be very effective for training generic text localisation systems. Still, the lack of realism (both in terms of positioning, and blending with the scene), diversity (in terms of text styles and scene backgrounds) and contextualisation of the text in the scene, have been limiting factors. More recent works aim to improve some of these aspects~\cite{chen2019cross, zhan2019ga, liao2019synthtext3d, long2020unrealtext}, or exploit instead real scene text data to do augmentation~\cite{gomez2019selective}, still, cannot replace the quality of real-world data.
	
	\subsection{Weakly supervised learning}
	The proposed method builds on top of pseudo-labelling techniques \cite{lee2013pseudo}, a simple strategy for semi-supervised learning where part of the data is fully labelled and Pseudo-Labels are created for unlabelled data as the class with the maximum predicted probability and further treated as true labels.

	Focusing on Chinese street view images, \cite{sun2019chinese} have rough location and transcription of some text instances annotated. An online proposal matching module is incorporated in the whole model. The main difference from our method is that they do not do any modification of the proposed regions.

	In \cite{Qin_2019_ICCV}, an existing OCR engine different from the one being trained is used to provide partial labels for one million unlabelled images. The partially labelled data is then used to train the recognition part of an E2E model, improving the results significantly. The method relies on a confidence threshold to filter out noisy labels while our method relies on weak annotations to minimize the risk of incorrect labelling.

	Focusing on text detection, \cite{qin2019curved} propose multiple approaches for unsupervised and weakly supervised learning. Their unsupervised approach simply relies on filtering out predictions with low confidence score.
	An improved approach requires weak annotations, where regions containing text are annotated and it is known that regions distant from the annotated ones do not contain any text, allowing for more accurate false positives filtering.
	Their last approach relies on rectangular bounding boxes as weak labels.
	
	\section{Datasets}

    We introduce the datasets used in our experiments.
	
	\textbf{Amazon Book Covers (ABC)} is a dataset created from more than 200,000 images downloaded from Amazon Books. The author and the title of each book serve as weak annotations. The same data were already used for genre prediction in \cite{iwana2016judging}.
	Some illustration images are shown in Figure \ref{fig:abc-examples}.
	
	\textbf{Uber-Text dataset (UT)} is one of the biggest datasets for text detection and recognition. It contains 117,969 images with 571,534 labelled text instances split into training, validation and test sets. Each set is divided into two subsets according to the image resolution - either 1K or 4K. The images were obtained through the Bing Maps Streetside program and come from 6 different cities in the US. The annotations are line-level. Most of the text regions form semantic units such as business names, street signs or street numbers. 
	The datasets contains a lot of not annotated text, some text regions are not annotated at all, while some readable text is labeled as unreadable \cite{zhang2017uber}.

	
		
	
	\textbf{MJSynth (MJ)} 
	contains almost 9M synthetically generated images of English words for text recognition.
	The text generation process performs the following steps: 
	Font rendering, border/shadow rendering, coloring, projective distortion, natural data blending and noise introduction \cite{jaderberg2016reading}. 
		\textbf{SynthText (ST)}
	is a synthetic dataset designed for scene-text detection, widely used for recognition, too. It has over 7M text instances in 8,000 images \cite{gupta2016synthetic}.
\textbf{Synthetic Multi-Language in Natural Scene Dataset (MLT)} contains 245,000 images in total with text instances in multiple scripts: Arabic, Bangla, Chinese, Japanese, Korean and Latin. The dataset was published in \cite{buvsta2018e2e} and the authors have adapted the framework of \cite{gupta2016synthetic}. A non-latin dictionary was used and it contains special, non-alpha-numeric characters.
	We only use the Latin script subset of the dataset, which contains 288,917 text instances in total \cite{buvsta2018e2e}.
	
	\textbf{IIIT 5K-word (IIIT)} is a collection of 5,000 cropped words from Google image search using queries
	which are likely to contain text \cite{MishraBMVC12}. The training set consists of 2,000 images, the remaining 3,000 form the test set.\textbf{Street View Text (SVT)} was collected from the Google Street View, providing annotators with a lexicon for each image, containing texts such as business names. Only the words from the lexicon were localised and provided with transcription, the rest of the text is ignored. There are 257 and 647 images of cropped words in the training and test sets~ \cite{wang2011end}.
	\textbf{Street View Text - Perspective (SVT-P)} is a dataset of 645 images collected from Google Street View focused on perspective projections \cite{Phan_2013_ICCV}.
	
	\textbf{ICDAR2003 (IC03)}, collected for the ICDAR 2003 Robust Reading competitions~\cite{lucas2003icdar}, has 258 training and 251 testing images with 1,156  and 1,110 annotated words respectively. 
	\textbf{ICDAR2013 (IC13)} is a dataset with 'focused text', the text being the main content of the image. It consists of a training set of 229 images with 848 words and a test set of 233 images with 1095 words.
	\cite{karatzas2013icdar}. In contrast,
		\textbf{ICDAR2015 (IC15)}, focuses on incidental scene-text - the images were not taken with text in mind. The training set contains 1,000 images (4,468 words) and the test contains 500 images (2,077 words) \cite{karatzas2015icdar}.
	
	
	\textbf{Total-Text (TT)} is a dataset of 1,555 scene images with 9,330
	annotated words. The images were collected with curved text in mind and the images often contain texts of different orientations \cite{CK2019}.
\textbf{CUTE80 (CT)} contains 80 images with 288 words, focusing on curved text \cite{risnumawan2014robust}.

	\section{PGT Generation}

	\label{sec:pgt_gen}

This section describes the pseudo ground truth (PGT) generation algorithm (PGT-GEN). The algorithm uses weakly annotated images and an existing end-to-end reading system (E2E). All the steps are executed independently for each image.
First, we define the E2E output and the structure of the weak annotations. Then we describe the algorithm and its components in detail. 

Given an image $I$, the output $O =~ \{(bb_1, tt_1), \dots (bb_{t}, tt_{t})\}$ of the end-to-end reading system 
is a set of $t$ text bounding box predictions and the corresponding text transcriptions.
The transcriptions $T = (tt_1, \dots, tt_{t})$ are strings (possibly containing spaces) and the bounding boxes are oriented rectangles.
It is necessary that the recognition output from a bounding box $bb$ can be obtained independently: $\text{REC}(I, bb) = tt$.

Each image is associated with a list of texts $ A =~ (t_1, t_2 \dots t_n ) $ where each text $ t_i = (w_1, w_2, \dots, w_m )$ is a non-empty ordered sequence of words.
The set of weak labels $ G = \bigcup^{n}_{i=1}g_i $ is obtained as a union of sets of k-grams, $k \in \{1, .. 5\}$. Each set of k-grams $g_i$ is formed by strings (consecutive words from $t_i$), sub-sequences of $t_i$ of length $k$ joined into a single string by the space character. 
In the simplest of cases, each text $t_i$ only consists of a single word but because the texts are assumed to be extracted automatically as metadata accompanying the images, it may even be multiple words that form a semantic unit --- a name of a product, its description, a business' name, contact information. These words are likely to appear in the image close to each other and get merged by the detector. 

For example, 
the texts could be
``Sherlock Holmes'' and ``221B Baker Street''. $A$ and $G$ would then be
\begin{gather*}
A = ( (\text{``Sherlock''}, \text{``Holmes''}), 
(\text{``221B''}, \text{``Baker''}, \text{``Street''})) \\
G= \{\text{``Sherlock''}, \text{``Holmes''},  \text{``Sherlock Holmes''}, 
\text{``221B''}, \\ \text{``Baker''}, \text{``Street''}, \text{``221B Baker''},  \text{``Baker Street''}, \\ \text{``221B Baker Streeet''}\}.
\end{gather*}

\subsection{PGT-GEN algorithm}

The PGT-GEN algorithm takes the image $I$, E2E output $O$ and the set of weak labels represented as k-grams $G$ as an input and outputs the PGT - a localized subset of $G$. 

{
\begin{algorithm}
\SetAlgoLined
\KwIn{$I, O, G$ }
\KwOut{$PGT$}
$P$ := AssignWeak($O$, $G$)\;
$PGT$ = $\{ \}$\; 
\ForEach{ $((bb, tt), g) \in \text{P}$}
{
	$(bb_f, tt_f)$ = FindOptimalBox($I, bb, tt, g$)\; 
	\If{{\rm IsPGT}($tt_f, g$)} 
{
	$PGT$ = $PGT$ $\cup \{(bb_f, g)\} $\;
}
	
}
\KwRet{$PGT$}
 \caption{PGT-GEN}
 \label{alg_pgt}
\end{algorithm}
}

\textbf{AssignWeak - Weak annotation assignment.} Each element from $O$ is assigned at most one weak annotation from $G$. 
We construct a directed bipartite graph $B_G = (V, E)$ between $O$ and $G$, thus $V = O \cup G$.
For each output $o \in O$, $o = (bb, tt)$ and weak annotation $g \in G$ it holds that
\begin{equation}
(o, g) \in E \iff  \text{dist}(tt, g) = \min_{i=1}^{|G|}\text{dist}(tt, g_i)
\end{equation}
\begin{equation}
 (g, o) \in E \iff \text{dist}(tt, g) = \min_{i=1}^{|T|}\text{dist}(tt_i, g)
\end{equation}
where dist is the Levenshtein distance.

Then, a set of proposals $P$ is created:
\begin{equation}
	P = \bigcup^{|O|}_{i=1}\text{Assign}(o_i, E) 
\end{equation}

\begin{equation}
\text{Assign}(o, E) =
\begin{cases}
\emptyset & \text{for }  W(o, E) = \emptyset \\
[W(o, E)]_R & \text{otherwise} \\
\end{cases}
\end{equation}

\begin{equation}
	W(o, E) = \{(o, g): (o, g) \in E \land (g, o) \in E  \land \text{match}(o, g) \}.
\end{equation}
We define $\text{\text{match}}((bb, tt), g) = \frac{\text{dist}(tt, g)} {{\max (\text{len}(tt), \text{len}(g))}} < 1$ to filter out completely irrelevant proposals
and $[.]_R$ selects an element from a set randomly. In most cases, $|W(o, E)|\in \{0, 1 \}$.

At this point, we could apply some simple filtering to the set of proposals $P$ instead of the edit distance guided neighbourhood search, for example, select
\begin{equation}
P' = \{p \in P, p = ((bb, tt), g)| dist(tt, g) = 0 \}
\end{equation}
 and then output 
 \begin{equation}
 \text{PGT} = \bigcup_{((bb, tt), g) \in P'}(bb, tt).
 \end{equation}
 This would be equivalent to selecting proposals where the predicted transcription was equivalent to the weak label text for PGT - we implement this version and compare it to the proposed one, showing the superiority of the proposed method.

\textbf{FindOptimalBox - Edit distance guided neighbourhood search.} 
For each proposal $((bb, tt), g) \in P$, 
we search for an optimal bounding box $bb_f$ which minimizes the Levenshtein distance between the recognized text $tt_f$ and $g$.

If $\text{dist}(tt, g) = 0$, we assume that $bb$ is already optimal and assign $bb_f = bb$.
If not, 
we predefine a set of new boxes in the neighbourhood of $bb$ and run the recognition on those in parallel, selecting one with minimal distance from $g$ for $bb_f$. The generation of the set of predefined boxes is explained in detail in Appendix \ref{app_bb_transform}.

We compute $	tt_f = \text{REC}(I, bb_f)$
and the normalized edit distance between $tt_f$ and $g$ as
$d =\frac{\text{dist}(tt_f, g)}{\max (\text{len}(tt_f), \text{len}(g))}$. 

Finally, we find out whether $(bb_f, tt_f)$ satisfies our requirements for being a PGT (\textbf{IsPGT}) as:
\begin{equation}
\text{IsPGT}(tt_f, g) =
\begin{cases}
\textrm{True} & \text{for }  d = 0 \lor \text{isClose}(d, tt_f, g) \\
\textrm{False} & \text{otherwise} \\
\end{cases}
\end{equation}
where 
$ \text{isClose}(d, tt_f, g) = (d < \theta \land |tt_f| > \lambda  \land tt_f^0 = g^0 \land tt_f^{-1} = g^{-1})$,
$s^0, s^{-1}$ are the first and the last characters of a string $s$.
The thresholds $\theta$, $\lambda$ are set to $ \theta = 0.35, \lambda = 4$.
The intuition behind the IsClose function is that even if the recognized text $tt_f$ and the assigned text $g$ are not identical, it is possible that there was simply an error in the recognition step. If the relative edit distance between two longer texts is low and the first and the last characters are the same, it is likely that $tt_f$ should actually be $g$. 
	
Examples of how the neighbourhood search aids the PGT generation process can be seen in Figure~\ref{fig:pgt_search}.

{
	 \setlength{\tabcolsep}{2pt}
	\begin{figure}
		\centering
		\begin{tabular}{cc}

			\includegraphics[trim={0 0 0.75cm 0},clip, width=0.23\textwidth]{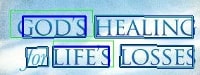} &
			\includegraphics[trim={0 0.5cm 0 0.5cm},clip, width=0.25\textwidth]{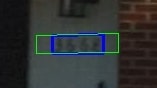} \\

		\end{tabular}
		
\vspace{-1ex}
		\caption{Improved localization of weak labels by the neighbourhood search - the detected bounding boxes (blue) and the transformed ones (green). 
		}
		
		\label{fig:pgt_search}
		    \vspace{-3ex}
	\end{figure}
	
}

	\section{End-to-end reading system}
	
        	In this section, the end-to-end reading system (E2E) used in our experiments is described. Separate models for detection and recognition are used.
	
	\subsection{Detection}
	
	For  text  detection,  we  adopt  TextSnake  \cite{long2018textsnake}.  It is based on a fully convolutional network -- U-net with VGG-16 \cite{simonyan2014very} backbone -- which estimates the geometry attributes of text instances.
	A text instance is described as a sequence of ordered, overlapping disks centered at symmetric axes. Each disk is associated with a potentially variable radius and orientation.
	
	The following values are predicted for each pixel: $tcl, tr, r$ and $\alpha$, corresponding to the text center line, text region, radius and angle.
	Thresholds varying on different datasets are applied to $tcl$ and $tr$ to obtain binary masks  $tcl_b$ and $tr_b$. 
	Focusing on straight text, we replace the proposed subsequent post-processing steps for text instance reconstruction with a method based on least squares fitting of the text center line points, which improves the orientation of the final bounding box.
	
	First we obtain the connected components $\texttt{CC}$ from $tr_b * tcl_b$.
	For each component $cc \in \texttt{CC}$, where $cc$ are all the component points, we estimate the bounding box $(c_x, c_y, w, h, \alpha)$ directly.
	
	The angle $\alpha$ is estimated by total least squares fitting of a line to the points of $cc$ shifted by the mean value of the coordinates $ m = (m_x, m_y)$ to the origin, using the slope of the best fitting line as the bounding box angle.
	
	The height $h$ is determined via the biggest radius predicted within the component: $h = \max(r[cc]) * 2 $.	
	To determine the width $w$, we project the shifted points onto the line with slope $\alpha$ passing through the origin and find the projected vectors with maximum norm in both directions, $p_{pos}$ and $p_{neg}$. The width is calculated as $w = |p_{pos} - p_{neg}| + h $. The extra $h$ is added to the width because during training, the $tcl$ is shrank by $\frac{h}{2}$ 
	(assuming the radius is constant, $\frac{h}{2}$, for straight text with rectangular ground truth). Afterwards, we shift $p_{pos}, p_{neg}$ back by $m$ and calculate the center of the bounding box as the middle point:
	$(c_x, c_y) = m + \frac{p_{pos} + p_{neg}}{2}$.
	
	
%
	
	\subsection{Recognition}
	
	We adopt the best performing architecture from \cite{baek2019wrong}, similar to STAR-net \cite{liu2016star} but with a different prediction mechanism.
	
	\subsubsection{Transformation}

	An input image $I$ is transformed into a rectified image $\tilde{I}$. It predicts the parameters of a thin-plate spline (TPS) transformation, a variant of spatial transformer network (STN) \cite{jaderberg2015spatial}. The whole module consists of a  localization network, a grid generator and a grid sampler. 
	
	We use grayscale images as the input and both the input and output dimension are fixed to 32 $\times$ 150 pixels.
	For more details, we refer the reader to \cite{shi2016robust, baek2019wrong, liu2016star}.

	\subsubsection{Feature Extraction}

	Given the rectified image $ \tilde{I} $, the feature extractor outputs a feature map
	\begin{equation}
 		V =\text{CNN}(\tilde{I}) = \{v_i\}, i = 1, \dots K 
	\end{equation}
    where $K=38$ is the number of columns in the output feature map ($512 \times 38 $).

	\subsubsection{Sequence modeling} a
	BiLSTM network \cite{graves2005bidirectional} creates contextual features from the visual features $ {v_i} $ and outputs  
	$ H = \text{Seq}(V) $. 
	We use a 2-layer BiLSTM.
	An $ i^{th}$ layer identifies two hidden states: forward $ h_i^{(t), f}$ and backward $ h_i^{(t), b}  \forall t$. A fully-connected layer between the two BiLSTM layers determines one hidden state, $ \hat{h}_t^{(i)}$, from  $ h_i^{(t), f}$ and $h_i^{(t), b}$. The dimension of the hidden states and the FC layer is 256.
	
	\subsubsection{Prediction}
	
	Finally, a single layer LSTM~\cite{hochreiter1997long} attention decoder produces the output sequence of characters 
	$ Y =~ y_1, y_2, \dots y_n$, 
$	y_t = \text{softmax}(W_o s_t + b_o),$
%
	\noindent where $W_o$ and $b_0$ are trainable parameters, and $s_t$ is the decoder LSTM hidden state at time $t$. The decoding stops when the {\tt(EOS)} symbol is emitted. The model is trained with the cross entropy loss function. For more details on the attention mechanism, please see \cite{Baek_2019_CVPR, cheng2017focusing, Qin_2019_ICCV}.
	
	This recognition model is used in all of our experiments and we will refer to it simply as $\text{OCR}$.

	\section{Experiments}
	
    	The pseudo ground truth (PGT) generation method was tested with two different sources of weakly annotated data, the Amazon book covers dataset (ABC) and the Uber-Text (UT) training set where we ignored 
	localization information.
	
	The detection part of E2E (TextSnake) was trained on a mix of SynthText, ICDAR2015 and Total-Text datasets. 
	The post-processing thresholds of TextSnake were set to $tr = 0.4$ and $tcl = 0.7$, which lead to the best PGT generation performance on a small subset of  UT and ABC images.
	We do not filter out the text marked as {\tt unreadable} or {\tt don't care} during training to maximize the use of available data. Detection recall is more important than precision for PGT generation --  the more words detected, the more potential pseudo-labelled examples are available.
	On the other hand, false positives are very unlikely to be matched against weak annotations,
	thus they have minimal impact, 
	besides slowing down the process. 
	
	The recognition part (OCR), which also serves as a baseline model ($\text{OCR}_b$) in our experiments, was trained on the ST (Synth-text), MJ (MjSynth) and MLT (Synthetic   Multi-Language   in   Natural   Scene)  datasets.
	The $\text{OCR}_b$ recognizes 70 characters -- letters, not distinguishing lower and upper case, digits and frequent special characters like punctuation, brackets, the {\tt(EOS)} symbol and the space. 
	
	Most recognition datasets provide word-level annotations, and thus {\tt space} is never part of the transcription. We included {\tt space} in the character set for three different reasons.
	First, if the model is capable of predicting spaces, it helps to guide the PGT generation process - a bounding box that is too wide leads to a space being predicted at the beginning or end of the transcription.
	Second, if the detector merges horizontally adjacent words,
	the recognizer often splits the text by recognizing a {\tt space} between the merged words. Third, it allows exploiting annotations that contain multiple words.
	
	Synthetic datasets used for training have word-level annotations and thus provide no training data for the {\tt space} character. We therefore 
	extended some of the bounding-boxes and included spaces at the beginning and end of the ground truth annotations.
	This produced a  model with a limited ability to recognize the {\tt space}.
	It was further improved during training on PGT,
	since it  contains multi-word texts.
	During evaluation, we strip any leading/trailing spaces from the predictions.
	
\begin{figure}
    \centering
        \includegraphics[trim = 15mm 0mm 0mm 0mm, clip,width=0.99\columnwidth]{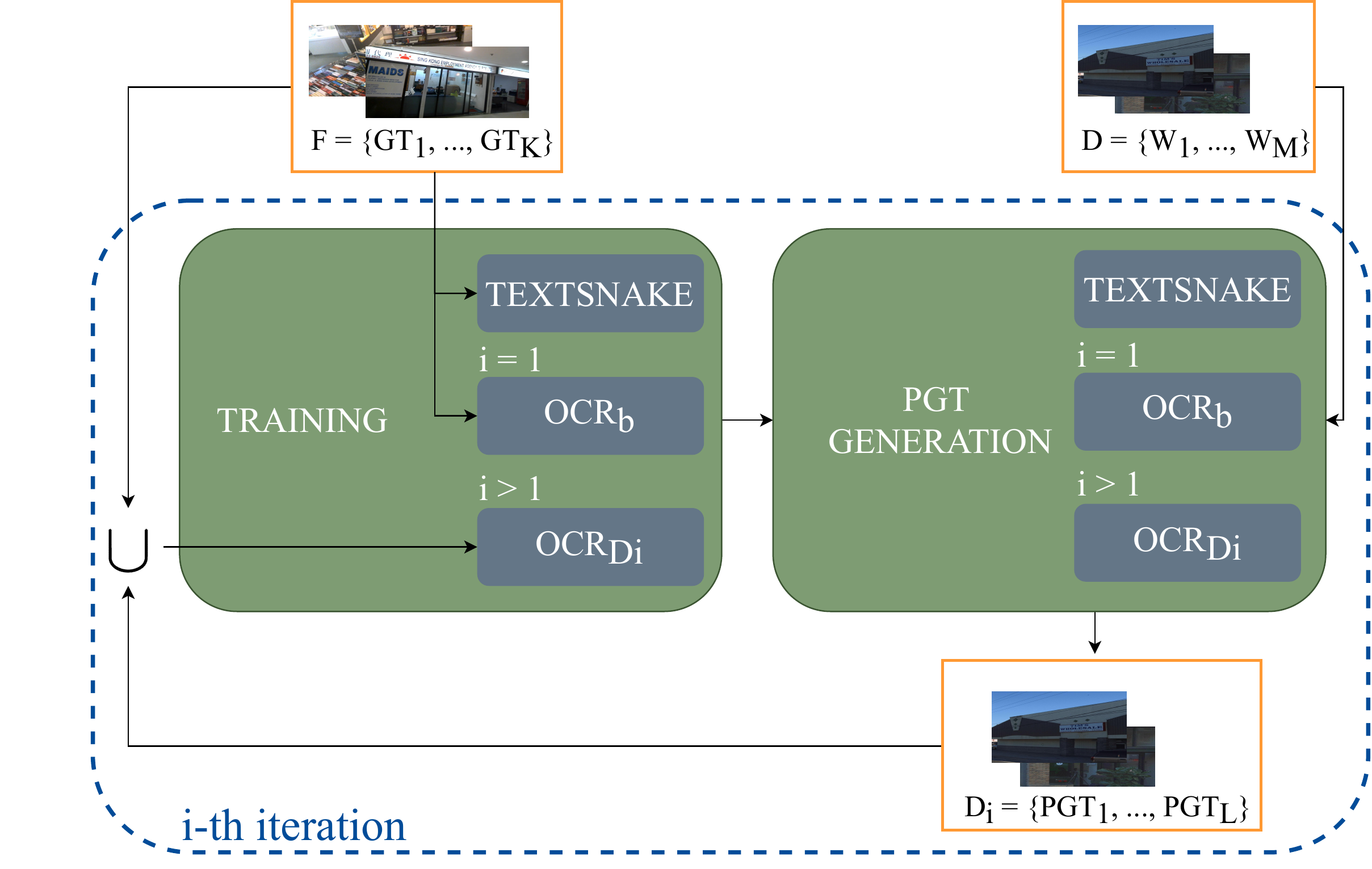}

    \caption{Training procedure. In the $i$-th iteration, $i > 1$, the outputs of the $ \text{OCR}_{\text{D}_{i-1}}$ and $\text{TextSnake}$ models are used to generate a pseudo ground truth (PGT) dataset $\text{D}_{i}$ from a weakly annotated dataset D. A new model $\text{OCR}_{\text{D}_{i}}$ is trained on the union of $\text{D}_i$ and fully annotated data F.
    In the first iteration, $\text{OCR}_\text{b}$ and $\text{TextSnake}$ are pretrained on F. }
    \label{fig:pgt_training}
        \vspace{-2ex}
\end{figure}

	The OCR processes images with a fixed resolution of $32 \times~ 150$.
	Input images are first resized isotropically to height 32. 
	If the width of the resized image is less than 150, the image is extended to the left and padded with zeros.
	If the width exceeds 150, 
	it is horizontally shrunk to 150 - only in this case the aspect ratio of the input images is not preserved.
	The procedure of training with PGT is explained in Figure \ref{fig:pgt_training}.

	\subsection{PGT from the Uber-Text dataset}
	The experiment evaluates the PGT method as an adaptation technique to the Uber-Text dataset domain. 
	The performance is also compared to fully-supervised and semi-supervised training in the UT domain.

   A reference method, $ \text{OCR}_{\text{UT}_{\text{F}}}$, is obtained by
	fully supervised training on  $\text{UT}_\text{F}$,  the set of 138,437 transcriptions and corresponding rectangular crops from the UT training set 
	that contain no unreadable characters in transcription. 
	The crops are the minimum area enclosing rectangles of the ground truth polygons.
	Appendix \ref{app:add_exps} contains details of a semi-supervised learning via pseudo-labelling experiment. The best performing model from this experiment is referred to as $ \text{OCR}_{\text{UT}_{\text{PL}}}$. 
	$ \text{OCR}_{\text{UT}_{\text{F}}}$ and $ \text{OCR}_{\text{UT}_{\text{PL}}}$, as well as other OCRs described in this section, are validated on a set of 5,000 random transcriptions from the UT validation set.
	
	For PGT generation, the whole UT training set is used. 
	To facilitate GPU computations, we split large (about 4K) images into 16 blocks ensuring no text instance is split and discard those  with no text. Such empty blocks are common since text instances are sparse in many images.
	Each of the original ground truth transcriptions  is a weak label in our experiment, the
	ground truth polygons are discarded.  The weak labels are transformed into a set of n-grams,  as explained in the PGT generation section. N-grams containing the {\tt *} symbol (unreadable or unknown characters) are discarded.
	
\begin{figure}
    \centering
    \includegraphics[width=0.95\columnwidth]{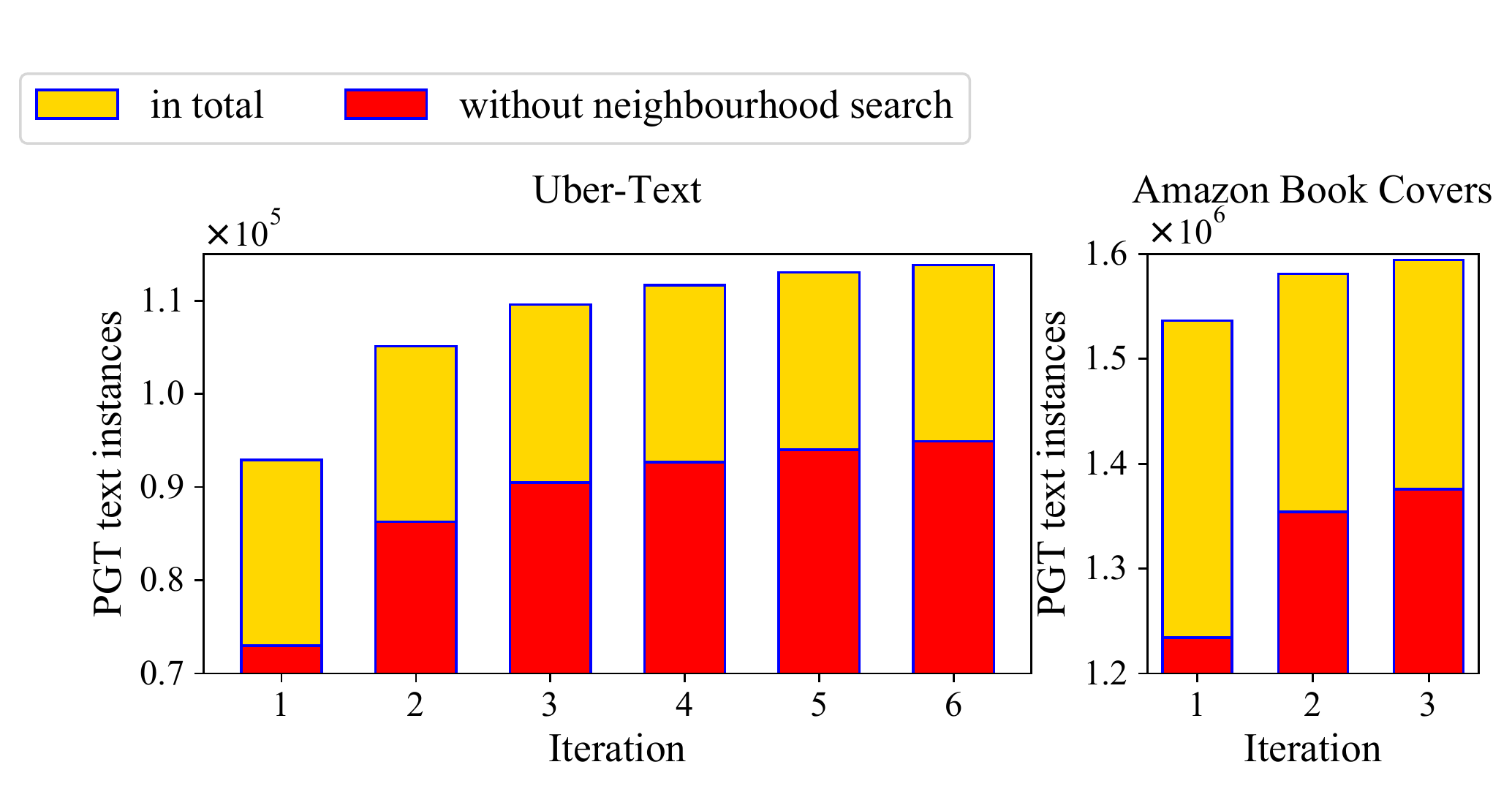}
    \caption{Number of PGT text instances on the UT and the ABC datasets. On UT, the text location information is ignored. }
    \label{fig:pgt_numbers}
        \vspace{-2ex}
\end{figure}

	{\it PGT generation and OCR training}. In the first iteration, 92,909 PGT text instances were obtained.
	The number of PGT text instances  increased in all iterations, reaching 113,810 texts after six iterations when the OCR performance stopped improving -- a summary is shown in Figure \ref{fig:pgt_numbers}.
	The recognition rate, calculated on 20,000 randomly selected transcriptions from the UT test set, increased in each iteration from the baseline 41.6 \% to 66.1~ \% in the sixth iteration.
	The accuracy of the fully supervised $\text{OCR}_{\text{UT}_\text{F}}$ and semi-supervised $\text{OCR}_{\text{UT}_\text{F}}$ is 78 \% and 45.7 \%, respectively. The PGT has reduced the gap between the baseline model and the fully-supervised one by 67 \%.
	The performance of PGT training is limited by the detector which was not trained on the new domain. Improving the detector may help to reduce the gap further.

    To test the contribution of the neighbourhood search and of allowing imperfect matches,  a dataset, denoted $\text{UT}_{\text{1}}'$, is created.  It contains only the detections that matched with 0 edit distance with some of the weak labels. The accuracy of the model trained with  $\text{UT}_{\text{1}}'$ is 2.7 \% lower, showing the importance of the additional retrieved PGT text.
    
\begin{figure}
    \centering
    \includegraphics[width=0.95\columnwidth]{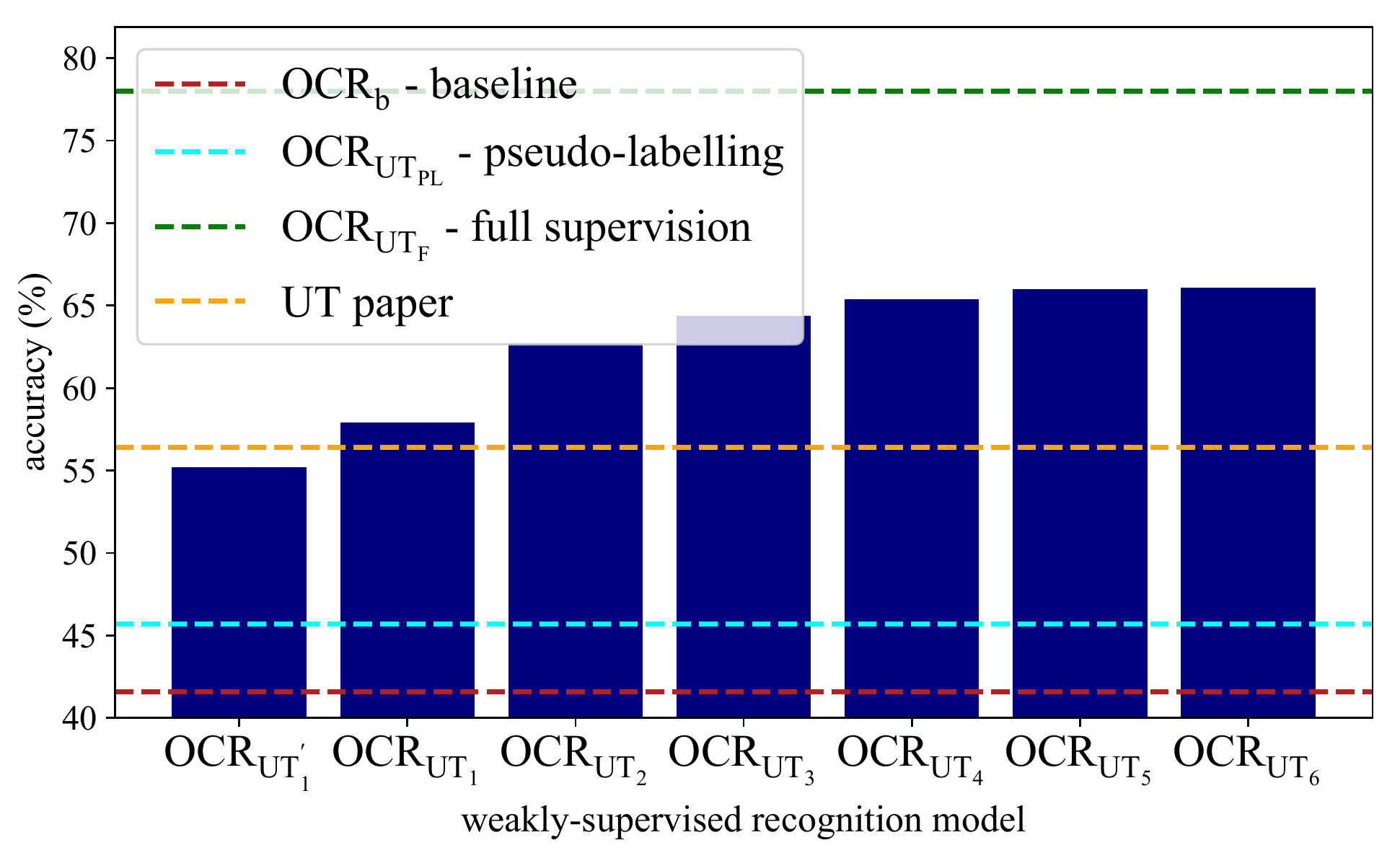}
    \caption{Recognition rates on the Uber-Text test set.
	The model trained in iteration $i$ is denoted as $\text{OCR}_{\text{UT}_i}$. $\text{UT}_\text{1}'$ is a subset of
	$\text{UT}_1$ obtained without
	the neighbourhood search. 
	UT paper is the model of \cite{zhang2017uber}. }
    \label{fig:ut_acc}
    \vspace{-2ex}
\end{figure}

    Figure \ref{fig:ut_acc} compares the accuracy of different models and Tables \ref{tab:mining} and \ref{tab:uber-test} summarize the different experiments.

	
	
	
	\subsection{PGT from book covers}
	
	The PGT is generated from the whole ABC dataset - over 200,000 images, using the author and the title of each book as weak label. The title often includes a subtitle - while the author and title are almost always present in the image, the subtitle is less common, or there may only be a part of it.
	

	We performed three iterations, again increasing the number of PGT generated in each of them. In the last one, $\text{ABC}_3$, a dataset of 1,594,333 cropped images of text is generated.
	Figure \ref{fig:pgt_numbers} contains a summary of the PGT generation results and Table \ref{tab:mining} contains a more detailed results.

    \subsection{PGT accuracy}
    \label{subsec:pgt_acc}
    We have analyzed the accuracy of the PGT method on $\text{UT}_6$ and $\text{ABC}_3$, 500 text instance crops from each.
    We have found 21 and 10 images with a wrong PGT in the ABC and the UT datasets, respectively. Those are images that do not contain text or images where some characters/punctuation in the PGT are wrong. 
    The majority of these errors are due to false positives/very blurred texts, leading to a prediction of a common word such as `the', `on' with low confidence and thus can be filtered out.
     For more details, see Appendix \ref{app_pgt_acc}.
	


	\subsection{Results on benchmark datasets}
	
	A recognition model trained with the PGT data from the previous experiments is evaluated on different domains using various commonly used recognition datasets - IIIT 5K-word (IIIT), Street View Text (SVT), ICDAR2003 (IC03), ICDAR2013 (IC13), ICDAR2015 (IC15), Street View Text - Perspective (SVT-P) and CUTE80 (CT). We evaluate on test set subsets commonly used by researches as identified in \cite{baek2019wrong}. All the models were validated on a union of the training sets of all the previously mentioned datasets. The reported metric is the percentage of correctly recognized words.
	
	To evaluate models trained on word-level data only, a test set of 20,000 images where each image only contains a single word, referred to as $\text{UT}_\text{W}$, is also created.
	
	We also evaluate on the $\text{UT}_\text{W}$ subset of the UT dataset but it was not included in the validation set.
	We remove any spaces from all the predictions and when evaluating on datasets with images that contain punctuation but the ground truth does not, we filter any non-alphanumeric characters out.
	
	Training with either $\text{UT}_\text{1}$ or $\text{ABC}_\text{1}$ generated in the first iterations of the PGT generation consistently improves the performance. On some datasets, UT boosts the accuracy more than ABC and vice versa.  Training with both leads to a superior performance on all evaluated datasets. The data from the last iterations, $\text{UT}_{\text{6}}$ and  $\text{ABC}_{\text{3}}$, further improve the accuracy with an average improvement of 3.7 \% relative to  $\text{OCR}_b$.
	
	The model trained with the $\text{UT}_{\text{1}}'$ and  $\text{ABC}_{\text{1}}'$ datasets is also evaluated. Those datasets are subsets of the $\text{UT}_{\text{1}}$ and  $\text{ABC}_{\text{1}}$ datasets that would have been obtained if no neighbourhood search or edit distance filtering was used. With the exception of IC03 dataset, this model's performance is always inferior to the model trained with all the data.
	
	The results also show that while the baseline model, trained on synthetic data only, performs well over a wide range of datasets, the performance on UT is rather poor - only 52.8 \% accuracy. This shows the challenging nature of the dataset, partially due to the presence of heavily blurred images and the high frequency of vertical/diagonal text direction.
	The summary of the experiments can be seen in Table \ref{tab:strb-all}.
	
	For comparison with other methods, we trained and evaluated our best performing model on alpha-numeric characters only. The baseline model is pretrained with MJ and ST and fine-tuned with the the $\text{UT}_{\text{6}}$ and  $\text{ABC}_{\text{3}}$. During evaluation, all images with unknown characters are filtered out.
	The boost in performance here is slightly lower, 3.3 \% on average.
     This single model achieves second-best performance on three different datasets with respect to the most recent state-of-the-art models. Note that state-of-the-art performance is achieved by different architectures trained on different data across the benchmarks, furthermore, the model of \cite{litman2020scatter} uses multiple different model configurations.

\begin{table*}[t]
	
	\centering

		\setlength{\tabcolsep}{5.5pt}
	\begin{tabular}{|l|S[table-format=2]S[table-format=2]S[table-format=2]S[table-format=2]S[table-format=2]|ccccccc|ccc|c|S[table-format=2.1]|}
		
		\hline
		\multirow{3}{*}{} & 
		\multicolumn{5}{c|}{Training dataset - \% in batch} &
		\multicolumn{7}{c|}{Evaluation on}  &
		\multicolumn{3}{c|}{Summary} & &
		
		\\
		
		\cline{2-18}
		& \multicolumn{3}{c|}{Full} & \multicolumn{2}{c|}{Weak}
		& IIIT & SVT & \multicolumn{1}{c}{IC03} &  \multicolumn{1}{c}{IC13} & \multicolumn{1}{c}{IC15} & SP & CT & \multicolumn{3}{c|}{$\Delta$} & $\text{UT}_\text{w}$ & $\Delta$   \\ 
		\cline{2-18}
		& MJ & ST & MLT & $\text{UT}$ & $\text{ABC}$ & 3000 & 647 & 867 & 1015 & 2077 & 645 & 288 & avg & min & max & 20 000 &
		\\ 
		\hline 
		$\text{OCR}_b$ & 45 & 45 & 10 & 0 & 0 &  89.8 & 86.7  & 94.3 & 91.2 & 68.5 & 77.2 & 72.3 & 0.0 & 0.0 & 0.0 & 52.8 & 0.0
		\\
		$ \text{OCR}_{\text{ABC}_1}$ & 30 & 30 & 10 & 0 & 30 & 92.8 & 88.9 & 94.8 & 93.0 & 71.0 & 77.5 & 73.6  & 1.7 & 0.3 & 3.0 & 53.9 & 1.1
		\\
		$ \text{OCR}_{\text{UT}_1}$ & 30 & 30 & 10 & 30 & 0 & 92.2 & 88.6 & 95.0 & \textbf{94.1} & 70.6 & 79.2 & 76.4 & 2.3 & 0.7 & 4.1 & 61.6 & 8.8
		\\
		$\text{OCR}_{\text{UT}_1 + \text{ABC}_1}$ & 20 & 20 & 10 & 20 & 30 & 93.0 & 89.2 & 95.2 & \textbf{94.1} & 71.6 & 79.2 & \textbf{77.8}  & 2.9 & 0.9 & 5.5 & 61.8 & 9.0
		\\
		$\text{OCR}_{\text{UT}_6 + \text{ABC}_3}$ & 20 & 20 & 10 & 20 & 30 & \textbf{93.5} & \textbf{90.7} & \textbf{95.5} & 94.0 & \textbf{74.6} & \textbf{80.1} & \textbf{77.8}  & 3.7 & 1.2 & 6.1 & 67.8 & 15.0
		 \\
		 \hline
		$\text{OCR}_{\text{UT}_{\text{1}}' + \text{ABC}_{\text{1}}'}$  & 20 & 20 & 10 & 20 & 30 & 91.4 & 88.1  & 95.5 & 93.9 & 69.5 & 77.1 & 74.0  &  1.4 & -0.1 & 2.7 & 59.1 & 6.3 \\
		\hline
	\end{tabular} 

\vspace{-1ex}
	\caption{Recognition rate on standard benchmarks, non-alphanumeric characters included. Validation was performed on the union of training sets, with the exception of the UT dataset. Average, min. and max. improvements relative to $\text{OCR}_b$ ($\Delta$). $\text{UT}_\text{w}$ results are not added to the summary since PGT was extracted from the UT domain and higher improvements are observed.}
	
	\label{tab:strb-all}
\end{table*}%

\begin{table*}[t]
	
	\centering
	
	\setlength{\tabcolsep}{5pt}
	\begin{tabular}{|l|S[table-format=2]S[table-format=2]S[table-format=2]S[table-format=2]|ccccccc|ccc|c|S[table-format=2.1]|}
		\hline
		\multirow{3}{*}{} & 
		\multicolumn{4}{c|}{Training dataset - \% in batch} &
		\multicolumn{7}{c|}{Evaluation on} & \multicolumn{3}{c|}{Summary}& & \\
		\cline{2-17}
		& \multicolumn{2}{c|}{Full} & \multicolumn{2}{c|}{Weak}
		& IIIT & SVT & \multicolumn{1}{c}{IC03} &  \multicolumn{1}{c}{IC13} & \multicolumn{1}{c}{IC15} & SP & CT &\multicolumn{3}{c|}{$\Delta$} & $\text{UT}_\text{W}$ & $\Delta$    \\ 
		\cline{2-17}
		& MJ & ST & $\text{UT}_\text{6}$ & $\text{ABC}_\text{3}$ & 3000 & 647 & 867 & 1015 & 1922 & 645 & 287 & avg & min & max & 18 956 &
		\\ 
		\hline 
		 $\text{OCR}_b$  & 50 & 50 & 0 & 0 & 87.6 & 88.6 & 94.5 & 91.9 & 75.2 & 78.9 & 73.2 & 0 & 0 & 0 & 54.9 & 0.0  \\
	 	 $\text{OCR}_{\text{UT}_6 + \text{ABC}_3}$  & 25 & 25 & 20 & 30 & \cellcolor{LimeGreen!20} 91.7 & \cellcolor{LimeGreen!80} \underline{91.8} & \cellcolor{LimeGreen!80} \underline{95.7} & \cellcolor{LimeGreen!40} 94.2 & \cellcolor{LimeGreen!40} \underline{80.0} & \cellcolor{LimeGreen!20} 82.5 & 77.4 & 3.3 & 1.2 & 4.8 & 68.9 & 14.0\\
	 	 \hline
 
 	 Published SOTA  & \multicolumn{4}{c|}{\multirow{4}{*}{ \makecell{SOTA methods were each \\ trained on \\ different datasets}}} & 95.3 & 92.7 & 96.6 & 96.4 & 82.8 & 87.0 & 88.5 & \multicolumn{3}{c}{} & \multicolumn{2}{c|}{}\\
	 References & \multicolumn{4}{c|}{} & \cite{liao2019mask} & \cite{litman2020scatter} & \cite{litman2020scatter} & \cite{lu2019master, liao2019mask} & \cite{litman2020scatter}  & \cite{litman2020scatter} & \cite{liao2019mask} & \multicolumn{3}{c}{} & \multicolumn{2}{c|}{}
\\
	\cline{1-1}
	\cline{6-12}

	 SOTA $2^{\text{nd}}$ & \multicolumn{4}{c|}{}  & \underline{94.4} & \underline{91.8} & 95.4 & \underline{94.7} & 78.7 & \underline{83.6} &  \underline{87.5} & \multicolumn{3}{c}{} & \multicolumn{2}{c|}{}\\
	 References & \multicolumn{4}{c|}{} & \cite{yang2019symmetry} & \cite{liao2019mask}  & \cite{shi2018aster} & \cite{litman2020scatter, shi2018aster}  & \cite{yang2019symmetry}  & \cite{yang2019symmetry} & \cite{litman2020scatter, yang2019symmetry} & \multicolumn{3}{c}{} & \multicolumn{2}{c|}{} \\

		 \hline
	\end{tabular} 

\vspace{-1ex}
	\caption{
		Recognition rate on standard benchmarks, non-alphanumeric characters excluded. Like many published methods, the model was pre-trained with MJ and ST datasets. Average, min. and max. increment relative to $\text{OCR}_b$ ($\Delta$). The \underline{second best} scores and our score within \colorbox{LimeGreen!80}{1 \%}, \colorbox{LimeGreen!40}{3 \%}, \colorbox{LimeGreen!20}{5 \%} from SOTA highlighted. 
	} \label{tab:strb-alphanum}
\vspace{-2ex}

\end{table*}%
	
	The results of our work show that training with automatically generated PGT from very different domains, such as born-digital documents, can significantly improve the performance of a recognition model over a wide range of scene-text datasets. Also, adding only a relatively small number of those images helps significantly, implying the variety of data is important.
	Some common characteristics of the images where the PGT data has improved the model's performance are blurred images, perspective distortions, artistic/handwritten fonts or occluded/cropped characters. Examples are shown in Figure \ref{tab:ocr_impr_examples} while images where the performance has deteriorated are shown in Figure \ref{tab:ocr_deter_examples}.

{
		 \setlength{\tabcolsep}{2pt}
	
	\begin{figure}
		\small

		\centering

		\begin{tabular}{m{2.3cm}m{2.3cm}m{2.3cm}}

		\includegraphics[width=2.3cm, height=0.8cm]{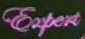}
		&
		\includegraphics[width=2.3cm, height=0.8cm]{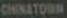}
		&
		\includegraphics[width=2.3cm, height=0.8cm]{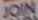}
		\\
		
		dopert & chantronn & jow \\
		$ \rightarrow$ expert & $ \rightarrow$  chinetown & $ \rightarrow$ join \\

	\includegraphics[width=2.3cm, height=0.8cm]{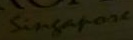}
		&
		\includegraphics[width=2.3cm, height=0.8cm]{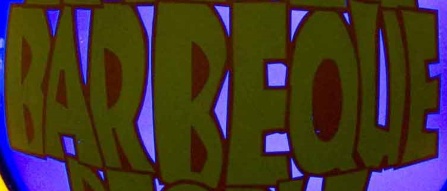}
		&
		\includegraphics[width=2.3cm, height=0.8cm]{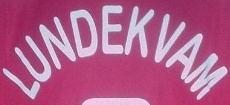}
		\\
		
		sireapost & brioffoole & underving \\
		$ \rightarrow$ singaport & $ \rightarrow$ barbeque & $ \rightarrow$ lundekvui \\

		
		
		

		\end{tabular}

\vspace{-1ex}
		\caption{Images with improved results after PGT  training.
	    The original $\text{OCR}_b$  and $\text{OCR}_{\text{UT}_6 + \text{ABC}_3}$ ($\rightarrow$) predictions.
		}
	
		\label{tab:ocr_impr_examples}
	\end{figure}

}

{
		 \setlength{\tabcolsep}{4pt}
	
	\begin{figure}[t]
		\small

		\centering

		\begin{tabular}{m{2.3cm}m{2.3cm}m{2.3cm}}

		\includegraphics[width=2.3cm, height=0.8cm]{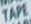}
		&
		\includegraphics[width=2.3cm, height=0.8cm]{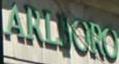}
		&
		\includegraphics[width=2.3cm, height=0.8cm]{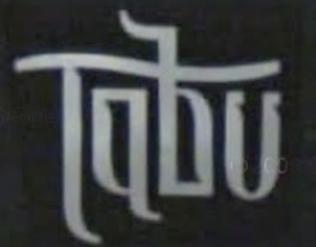}
		\\
		
		tape & arlboro & tqbu \\
		$ \rightarrow$ tapl & $ \rightarrow$  arljoro & $ \rightarrow$ tqlu \\
		
		\includegraphics[width=2.3cm,  height=0.8cm]{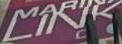}
		&
		\includegraphics[width=2.3cm,  height=0.8cm]{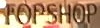}
		&
		\includegraphics[width=2.3cm,  height=0.8cm]{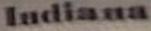}
		\\
		
		mink & topshop & indiana \\
		$ \rightarrow$ mark & $ \rightarrow$  forshop & $ \rightarrow$ ludiana \\
			
		\end{tabular}

\vspace{-2ex}
		\caption{Images with worse results after  PGT training. The original $\text{OCR}_b$ and $\text{OCR}_{\text{UT}_6 + \text{ABC}_3}$
		($\rightarrow$) predictions.
		}
	\vspace{-2ex}

		\label{tab:ocr_deter_examples}
	\end{figure}

}

	\section{Conclusion}
	We proposed a PGT generation method  and applied it to two different sources of weak annotations. As our baseline model, we chose the best-performing architecture with publicly available code-base \cite{baek2019wrong}. Training with PGT without any architecture changes consistently improved the recognition accuracy both on images from the same domain and across different benchmark datasets and thus different domains. 
	Our method is architecture-agnostic and thus can be applied to improve the performance of other architectures as well.

	

	\bibliographystyle{IEEEtran}
	\bibliography{IEEEabrv,main}

\begin{thebibliography}{10}
\providecommand{\url}[1]{#1}
\csname url@samestyle\endcsname
\providecommand{\newblock}{\relax}
\providecommand{\bibinfo}[2]{#2}
\providecommand{\BIBentrySTDinterwordspacing}{\spaceskip=0pt\relax}
\providecommand{\BIBentryALTinterwordstretchfactor}{4}
\providecommand{\BIBentryALTinterwordspacing}{\spaceskip=\fontdimen2\font plus
\BIBentryALTinterwordstretchfactor\fontdimen3\font minus
  \fontdimen4\font\relax}
\providecommand{\BIBforeignlanguage}[2]{{%
\expandafter\ifx\csname l@#1\endcsname\relax
\typeout{** WARNING: IEEEtran.bst: No hyphenation pattern has been}%
\typeout{** loaded for the language `#1'. Using the pattern for}%
\typeout{** the default language instead.}%
\else
\language=\csname l@#1\endcsname
\fi
#2}}
\providecommand{\BIBdecl}{\relax}
\BIBdecl

\bibitem{veit2016coco}
Veit \emph{et~al.}, ``Coco-text: Dataset and benchmark for text detection and
  recognition in natural images,'' \emph{arXiv:1601.07140}, 2016.

\bibitem{chen2019cross}
Chen \emph{et~al.}, ``Cross-domain scene text detection via pixel and
  image-level adaptation,'' in \emph{NIPS}.\hskip 1em plus 0.5em minus
  0.4em\relax Springer, 2019, pp. 135--143.

\bibitem{gomez2019selective}
R.~{Gomez} \emph{et~al.}, ``Selective style transfer for text,'' in
  \emph{ICDAR}, 2019, pp. 805--812.

\bibitem{zhan2019ga}
Zhan \emph{et~al.}, ``Ga-dan: Geometry-aware domain adaptation network for
  scene text detection and recognition,'' in \emph{ICCV}, 2019, pp. 9105--9115.

\bibitem{gupta2016synthetic}
Gupta \emph{et~al.}, ``Synthetic data for text localisation in natural
  images,'' in \emph{CVPR}, 2016, pp. 2315--2324.

\bibitem{liao2019synthtext3d}
Liao \emph{et~al.}, ``Synthtext3d: Synthesizing scene text images from 3d
  virtual worlds,'' \emph{arXiv:1907.06007}, 2019.

\bibitem{long2020unrealtext}
S.~Long and C.~Yao, ``Unrealtext: Synthesizing realistic scene text images from
  the unreal world,'' \emph{arXiv:2003.10608}, 2020.

\bibitem{Baek_2019_CVPR}
Baek \emph{et~al.}, ``Character region awareness for text detection,'' in
  \emph{CVPR}, June 2019.

\bibitem{Sun_2019_ICCV}
Sun \emph{et~al.}, ``Chinese street view text: Large-scale chinese text reading
  with partially supervised learning,'' in \emph{ICCV}, October 2019.

\bibitem{qin2019curved}
Qin \emph{et~al.}, ``Curved text detection in natural scene images with
  semi-and weakly-supervised learning,'' \emph{arXiv:1908.09990}, 2019.

\bibitem{zhang2017uber}
Zhang \emph{et~al.}, ``Uber-text: A large-scale dataset for optical character
  recognition from street-level imagery,'' in \emph{Scene Understanding
  Workshop-CVPR}, 2017.

\bibitem{epshtein2010detecting}
Epshtein \emph{et~al.}, ``Detecting text in natural scenes with stroke width
  transform,'' in \emph{CVPR}.\hskip 1em plus 0.5em minus 0.4em\relax IEEE,
  2010, pp. 2963--2970.

\bibitem{neumann2010method}
L.~Neumann and J.~Matas, ``A method for text localization and recognition in
  real-world images,'' in \emph{ACCV}.\hskip 1em plus 0.5em minus 0.4em\relax
  Springer, 2010, pp. 770--783.

\bibitem{jaderberg2016reading}
Jaderberg \emph{et~al.}, ``Reading text in the wild with convolutional neural
  networks,'' \emph{IJCV}, vol. 116, no.~1, pp. 1--20, 2016.

\bibitem{liao2019mask}
Liao \emph{et~al.}, ``Mask textspotter: An end-to-end trainable neural network
  for spotting text with arbitrary shapes,'' \emph{IEEE t. PAMI}, 2019.

\bibitem{long2018textsnake}
Long \emph{et~al.}, ``Textsnake: A flexible representation for detecting text
  of arbitrary shapes,'' in \emph{ECCV}, 2018, pp. 20--36.

\bibitem{jaderberg2015spatial}
Jaderberg \emph{et~al.}, ``Spatial transformer networks,'' in \emph{NIPS},
  2015, pp. 2017--2025.

\bibitem{simonyan2014very}
K.~Simonyan and A.~Zisserman, ``Very deep convolutional networks for
  large-scale image recognition,'' \emph{arXiv:1409.1556}, 2014.

\bibitem{he2016deep}
He \emph{et~al.}, ``Deep residual learning for image recognition,'' in
  \emph{CVPR}, 2016, pp. 770--778.

\bibitem{Graves2006ConnectionistTC}
Graves \emph{et~al.}, ``Connectionist temporal classification: labelling
  unsegmented sequence data with recurrent neural networks,'' in \emph{ICML},
  2006.

\bibitem{cheng2017focusing}
Cheng \emph{et~al.}, ``Focusing attention: Towards accurate text recognition in
  natural images,'' in \emph{ICCV}, 2017, pp. 5076--5084.

\bibitem{Liu_2018_CVPR}
Liu \emph{et~al.}, ``Fots: Fast oriented text spotting with a unified
  network,'' in \emph{CVPR}, June 2018.

\bibitem{buvsta2018e2e}
Bu{\v{s}}ta \emph{et~al.}, ``E2e-mlt-an unconstrained end-to-end method for
  multi-language scene text,'' in \emph{ACCV}.\hskip 1em plus 0.5em minus
  0.4em\relax Springer, 2018, pp. 127--143.

\bibitem{Qin_2019_ICCV}
Qin \emph{et~al.}, ``Towards unconstrained end-to-end text spotting,'' in
  \emph{ICCV}, October 2019.

\bibitem{lee2013pseudo}
D.-H. Lee, ``Pseudo-label: The simple and efficient semi-supervised learning
  method for deep neural networks,'' in \emph{Workshop on challenges in
  representation learning, ICML}, vol.~3, 2013, p.~2.

\bibitem{sun2019chinese}
Sun \emph{et~al.}, ``Chinese street view text: Large-scale chinese text reading
  with partially supervised learning,'' in \emph{ICCV}, 2019, pp. 9086--9095.

\bibitem{iwana2016judging}
Iwana \emph{et~al.}, ``Judging a book by its cover,'' \emph{arXiv:1610.09204},
  2016.

\bibitem{MishraBMVC12}
Mishra \emph{et~al.}, ``Scene text recognition using higher order language
  priors,'' in \emph{BMVC}, 2012.

\bibitem{wang2011end}
Wang \emph{et~al.}, ``End-to-end scene text recognition,'' in
  \emph{ICCV}.\hskip 1em plus 0.5em minus 0.4em\relax IEEE, 2011, pp.
  1457--1464.

\bibitem{Phan_2013_ICCV}
Trung \emph{et~al.}, ``Recognizing text with perspective distortion in natural
  scenes,'' in \emph{ICCV}, December 2013.

\bibitem{lucas2003icdar}
Lucas \emph{et~al.}, ``Icdar 2003 robust reading competitions,'' in
  \emph{ICDAR}.\hskip 1em plus 0.5em minus 0.4em\relax Citeseer, 2003, pp.
  682--687.

\bibitem{karatzas2013icdar}
Karatzas \emph{et~al.}, ``Icdar 2013 robust reading competition,'' in
  \emph{ICDAR}.\hskip 1em plus 0.5em minus 0.4em\relax IEEE, 2013, pp.
  1484--1493.

\bibitem{karatzas2015icdar}
{Karatzas} \emph{et~al.}, ``Icdar 2015 competition on robust reading,'' in
  \emph{ICDAR}.\hskip 1em plus 0.5em minus 0.4em\relax IEEE, 2015, pp.
  1156--1160.

\bibitem{CK2019}
C.~K. Ch’ng \emph{et~al.}, ``Total-text: Towards orientation robustness in
  scene text detection,'' \emph{IJDAR}, vol.~23, pp. 31--52, 2020.

\bibitem{risnumawan2014robust}
Risnumawan \emph{et~al.}, ``A robust arbitrary text detection system for
  natural scene images,'' \emph{Expert Systems with Applications}, vol.~41,
  no.~18, pp. 8027--8048, 2014.

\bibitem{baek2019wrong}
Baek \emph{et~al.}, ``What is wrong with scene text recognition model
  comparisons? dataset and model analysis,'' \emph{arXiv:1904.01906}, 2019.

\bibitem{liu2016star}
Liu \emph{et~al.}, ``Star-net: A spatial attention residue network for scene
  text recognition.'' in \emph{BMVC}, vol.~2, 2016, p.~7.

\bibitem{shi2016robust}
Shi \emph{et~al.}, ``Robust scene text recognition with automatic
  rectification,'' in \emph{CVPR}, 2016, pp. 4168--4176.

\bibitem{graves2005bidirectional}
Graves \emph{et~al.}, ``Bidirectional lstm networks for improved phoneme
  classification and recognition,'' in \emph{ICANN}, 2005, pp. 799--804.

\bibitem{hochreiter1997long}
S.~Hochreiter and J.~Schmidhuber, ``Long short-term memory,'' \emph{Neural
  computation}, vol.~9, no.~8, pp. 1735--1780, 1997.

\bibitem{litman2020scatter}
Litman \emph{et~al.}, ``Scatter: Selective context attentional scene text
  recognizer,'' \emph{arXiv:2003.11288}, 2020.

\bibitem{lu2019master}
Lu \emph{et~al.}, ``Master: Multi-aspect non-local network for scene text
  recognition,'' \emph{arXiv:1910.02562}, 2019.

\bibitem{yang2019symmetry}
Yang \emph{et~al.}, ``Symmetry-constrained rectification network for scene text
  recognition,'' in \emph{ICCV}, 2019, pp. 9147--9156.

\bibitem{shi2018aster}
Shi \emph{et~al.}, ``Aster: An attentional scene text recognizer with flexible
  rectification,'' \emph{TPAMI}, vol.~41, no.~9, pp. 2035--2048, 2018.

\end{thebibliography}

	\clearpage
	\appendices
    \section{Bounding box transformations}
\label{app_bb_transform}
We define the following transformations to generate a set of bounding boxes in the neighbourhood of an input bounding box: Extending/shrinking the bounding box on the left/right/top. Angle modification and bottom extension/shrinkage were also considered but the benefits were insignificant. 
To keep the computational cost reasonable, 
we also assume that changes to the left side of the bounding box
do not influence the recognition of the characters 
on the right side and vice versa, the optimization on each side is done independently.
Horizontally, we extend/shrink the box with width $w$ and height $h$ in each direction by up to $c$ characters, the character length being estimated as the average character length $ch_{avg} = \frac{w}{|text|}$. On the top, we extend by up to $\frac{h}{\beta}$ and shrink by up to $\frac{h}{\gamma}$.

Each of the transformed bounding boxes can be characterized by three integer parameters relative to the original bounding box - $(t, l, r)$ - defining the extension/shrink (distinguished by the sign) by $t, l, r$ units on top/left/right, where the horizontal unit is $\frac{ch_{avg}}{\delta}$ and the vertical unit is $\frac{h}{\kappa}$. Consequently, $l, r \in [-c \cdot \delta; c \cdot \delta]$ and $ t \in [-\frac{\kappa}{\gamma}; 2 \frac{\kappa}{\beta}]$. The transformed bounding boxes that exceed the image or do not overlap with the original one are immediately discarded.

To obtain the final bounding box $bb_f$ characterized by $ (t_f, l_f, r_f)$, we find an optimal bounding box in both directions (left and right). For each direction, we find the set of boxes $B = \{ (t_i, l_i, r_i) \}_{i=1 \dots n} $ that result in the lowest edit distance from $g$. The sets $ T = \bigcup_{(t_i, l_i, r_i) \in B} t_i$, $ L = \bigcup_{(t_i, l_i, r_i) \in B} l_i$ and $ R = \bigcup_{(t_i, l_i, r_i) \in B} r_i$ are created.

From the boxes transformed in the left direction, we obtain 
\begin{equation}
t_l = \min T
\end{equation}
and
\begin{equation}
l_f = \frac{\min L + \min(\max L, o +\min L ) }{2}
\end{equation}
Analogously, for the right direction, we obtain
\begin{equation}
t_r = \min T
\end{equation}
and
\begin{equation}
r_f =\frac{\min R + \min(\max R, o +\min R ) }{2} 
\end{equation}
We set
\begin{equation}
t_f = \max(t_r, t_l)
\end{equation}

In our experiments, the constants were assigned as follows:
$c=7$,
$\beta = 2$,
$\gamma = 4$,
$\delta = 4$,
$\kappa = 4$,
$o = 8$.
All of these but $o$ control how many bounding boxes will be generated and the trade-off between precision of the box and computation time.
They were selected ensuring all the generated boxes for one PGT proposal can be run in a single batch on GPU and by observing qualitative results. 
The $o = 8$ means that $tt_f$ won't be more than one average character length wider on the right/left than the smallest bounding box that gives the correct recognition output.
    \section{Additional experimental results}

\begin{table}[t]
	
	\centering

	\begin{tabular}{cc|r|r|r}
& Iteration / Mined: & with neigh. s. & w/o neigh. s. &  $\Delta$  \\
\hline
\parbox[t]{2mm}{\multirow{6}{*}{\rotatebox[origin=c]{90}{UT}}}
& 1 & 92,909 & 72,990 & 19,919  \\
& 2 & 105,126 & 86,295 & 18,831 \\
& 3 & 109,557 & 90,480 & 19,077 \\
& 4 & 111,663 & 92,660 & 19,003 \\
& 5 & 113,046 &  93,994 & 19,052 \\
& 6 & 113,810 &  94,890 & 18,920 \\
\hline
\parbox[t]{2mm}{\multirow{3}{*}{\rotatebox[origin=c]{90}{ABC}}}
& 1 & 1,536,583 & 1,234,219 & 302,364 \\
& 2 & 1,581,109 & 1,354,219 & 226,890 \\
& 3 & 1,594,333 & 1,375,571 & 218,762 \\
\hline

	\end{tabular} 
	
\vspace{-1ex}
	\caption{PGT generation on the Uber-Text training dataset where text location information is ignored. 
	The number of captions generated in iterations 1 to 6,  
	with and without  the neighbourhood search, and the difference $\Delta$.}
	\label{tab:mining}
\end{table}%

\begin{table}[t]

	\centering
		\begin{tabular}{l|S[table-format=2]|S[table-format=2]|S[table-format=2]|S[table-format=2]|c|c|c}
			\hline 
			& \multicolumn{5}{c|}{Training dataset - \% in batch} & \multicolumn{2}{c}{}  \\ 
			\cline{2-6}
			& \multicolumn{4}{c|}{Full} & \multicolumn{1}{c|}{Weak} & \multicolumn{2}{c}{}  \\
			\cline{2-8}
			& MJ & ST & MLT &  $\text{UT}_\text{F}$ &  UT (30) & Acc. & NED  \\ 
			\hline 
			\cite{zhang2017uber} &  &  &  &  & & 56.4 & \\
			$\text{OCR}_b$ & 45 & 45 & 10 & 0 & - & 41.6 & 35.2 \\	
			\hline
			$ \text{OCR}_{\text{UT}_\text{1}'}$  & 30 & 30 & 10 & 0 & $\text{UT}_\text{1}'$ & 55.2 & 28.0 \\
			$ \text{OCR}_{\text{UT}_1}$  & 30 & 30 & 10 & 0 & $\text{UT}_{\text{1}}$ & 57.9 & 26.2 \\
			$ \text{OCR}_{\text{UT}_2}$  & 30 & 30 & 10 & 0 & $\text{UT}_{\text{2}}$ & 62.7 & 23.3 \\
			$ \text{OCR}_{\text{UT}_3}$  & 30 & 30 & 10 & 0 & $\text{UT}_{\text{3}}$ & 64.4 & 22.5 \\
			$ \text{OCR}_{\text{UT}_4}$  & 30 & 30 & 10 & 0 & $\text{UT}_{\text{4}}$ & 65.4 & 21.5 \\
			$ \text{OCR}_{\text{UT}_5}$  & 30 & 30 & 10 & 0 & $\text{UT}_{\text{5}}$ & 66.0 & 21.1 \\
			$ \text{OCR}_{\text{UT}_6}$  & 30 & 30 & 10 & 0 & $\text{UT}_{\text{6}}$ & 66.1 & 21.2 \\
			\hline
			$ \text{OCR}_{\text{UT}_{\text{F}}}$  & 30 & 30 & 10 & 30 & - & 78.0 & 10.0 \\
	    	$ \text{OCR}_{\text{UT}_{\text{PL}99}}$  & 30 & 30 & 10 & 30 & 	$\text{UT}_{\text{PL}99}$ & 44.5 & 33.5 \\
			$ \text{OCR}_{\text{UT}_{\text{PL}90}}$  & 30 & 30 & 10 & 30 & $\text{UT}_{\text{PL}90}$ & 45.7 & 31.7 \\
			$ \text{OCR}_{\text{UT}_{\text{PL}80}}$  & 30 & 30 & 10 & 30 & $\text{UT}_{\text{PL}80}$ & 44.8 & 32.1 \\
			$ \text{OCR}_{\text{UT}_{\text{PL}50}}$  & 30 & 30 & 10 & 30 & $\text{UT}_{\text{PL}50}$ & 44.4 & 33.1 \\
			\hline
		\end{tabular}

\vspace{-1ex}
	\caption{Recognition rates and normalized edit distance (acc., NED) on the Uber-Text test set.
	The data obtained from the $i^{th}$ iteration
	of the PGT generation is denoted as $\text{UT}_i$. 
	$\text{UT}_\text{F}$ and  $\text{UT}_{\text{PL}t}$ are the fully annotated  and pseudo-labelled (with a threshold $t$) datasets, respectively. $\text{UT}_\text{1}'$ is a subset of
	$\text{UT}_1$ obtained without
	the neighbourhood search. 
	}
	\label{tab:uber-test}
	
\end{table}

\begin{table}[t]
	
	\centering

	\begin{tabular}{c|c}
Confidence threshold & Mined \\
\hline
0.99 & 100,242 \\
0.90 & 137,005 \\
0.80 & 154,560 \\
0.50 & 204,549 \\
\hline
	\end{tabular} 
\vspace{-1ex}
	\caption{The number of PGt data generated in the semi-supervised setup with different confidence thresholds. There were 365,200 detections in total.}
	\label{tab:mining_uber}
	\vspace{-2ex}
\end{table}%

\begin{table*}[b!]
	
	\centering
		\setlength{\tabcolsep}{4pt}
	\begin{tabular}{|l|S[table-format=2]S[table-format=2]S[table-format=2]S[table-format=2]S[table-format=2]|ccccccc|ccc|c|S[table-format=2.1]|}
		
		\hline
		\multirow{3}{*}{} & 
		\multicolumn{5}{c|}{Training dataset - \% in batch} &
		\multicolumn{7}{c|}{Evaluation on}  &
		\multicolumn{3}{c|}{Summary} & &
		
		\\
		
		\cline{2-18}
		& \multicolumn{3}{c|}{Full} & \multicolumn{2}{c|}{Weak}
		& IIIT & SVT & \multicolumn{1}{c}{IC03} &  \multicolumn{1}{c}{IC13} & \multicolumn{1}{c}{IC15} & SP & CT & \multicolumn{3}{c|}{$\Delta$} & $\text{UT}_\text{w}$ & $\Delta$   \\ 
		\cline{2-18}
		& MJ & ST & MLT & $\text{UT}$ & $\text{ABC}$ & 3000 & 647 & 867 & 1015 & 2077 & 645 & 288 & avg & min & max & 20 000 &
		\\ 
		\hline 
		$\text{OCR}_b$ & 45 & 45 & 10 & 0 & 0 &  89.8 & 86.7  & 94.3 & 91.2 & 68.5 & 77.2 & 72.3 & 0.0 & 0.0 & 0.0 & 52.8 & 0.0
		\\
		$ \text{OCR}_{\text{UT}_1}$ & 30 & 30 & 10 & 30 & 0 & \textbf{91.6} & \textbf{88.7} & \textbf{94.7} & \textbf{94.0} & \textbf{71.3} & \textbf{79.4} & \textbf{74.7} & 2.1 & 0.4 & 2.8 & \textbf{62.5} & 9.7
		\\
        $ \text{OCR}_{\text{UT}_{\text{PL}99}}$ & 30 & 30 & 10 & 30 & 0 & 91.0 & 87.5 & 94.1 & 93.6 & 68.7 & 76.4 & 71.2 & 0.4 & -1.1 & 2.4 & 55.9 & 3.1
        \\
        $ \text{OCR}_{\text{UT}_{\text{PL}90}}$ & 30 & 30 & 10 & 30 & 0 & 90.6 & 87.2 & 94.0 & 92.5 & 68.5 & 77.8 & 71.2 & 0.3 & -1.1 & 1.3 & 57.5 & 4.7
        \\
        $ \text{OCR}_{\text{UT}_{\text{PL}80}}$ & 30 & 30 & 10 & 30 & 0 & 90.4 & 87.0 & 94.2 & 93.2 & 68.5 & 77.2 & 73.3 & 0.5 & -0.1 & 2.0 & 56.7 & 3.9
        \\
        $ \text{OCR}_{\text{UT}_{\text{PL}50}}$ & 30 & 30 & 10 & 30 & 0 &  90.9	& 88.4	& 94.3	& 93.4	& 69.1	& 76.9 & 72.9	& 0.8 & -0.3 & 2.2 & 56.1 & 3.3
        \\
		\hline
	\end{tabular}

\vspace{-1ex}
	\caption{Results of $ \text{OCR}_{\text{UT}_1}$ are slightly different than the previous table because the models here were validated on uber-text.}
	
	\label{tab:strb-all-semi}
\end{table*}%

\label{app:add_exps}
In this section, we provide additional experimental results in detail.

The number of text instances located in each iterations of the PGT generation on both the UT and ABC datasets is reported in Table \ref{tab:mining}.

The results of different models, tested and validated on the UT dataset, are reported in Table \ref{tab:uber-test}. For each model, we provide the per-batch percentage of examples from each dataset during training,the accuracy and the normalized edit distance. 

\subsection{Semi-supervised learning via pseudo-labelling}
For comparison with prior work \cite{qin2019curved, Qin_2019_ICCV}, we implement the pseudo-labelling \cite{lee2013pseudo} approach to semi-supervised learning with confidence-tresholding of the pseudo-labels. Pre-trained models, in our case, $\text{OCR}_{\text{b}}$ and $\text{TextSnake}$, are used to generate predictions on a set of unlabelled data - the UT training set (ignoring the existing ground truth). Only the predictions with a recognition confidence above a thresholds $t$ are kept to retrain the recognition model. 

We implement the method with different thresholds, $t \in~ \{99, 90, 80, 50\}$, the results are reported in Table \ref{tab:uber-test}. The best performing model is also included in Figure \ref{fig:ut_acc}. The accuracy of this model is 12.2 \% lower then the accuracy in the first iteration using our proposed method, $\text{OCR}_{\text{UT}_1}$. 

\label{app:semi-sup}

    \section{PGT accuracy}
    \label{app_pgt_acc}
Different kind of errors and ambiguities can occur in the resulting PGT data. We report their numbers in Table \ref{tab:pgt_error_freqs}, computed on 500 sample crops from the UT dataset and 500 samples from the ABC dataset. 

The wrong text and not text errors are the most harmful ones. However, employing recognition confidence based filtering, these can be reduced significantly. This is because these errors mostly occur for very blurred texts or images with no text, where the network predicts a short, common word such as `the', `in', `on', `at' with low confidence.

The error classification is not always clear, some may overlap.

{
\begin{table*}
    \centering
    \begin{tabular}{l|rr|cc}
    Counted on 500 sample crops: & ABC & UT & PGT & Crop \\
    \hline
    
     wrong text & 19 & 6 & cancer &	\includegraphics[height=0.6cm, width=2cm]{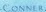} \\
    
    not text & 1 & 2 & the &
    \includegraphics[height=0.6cm, width=1.5cm]{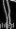} \\
    
    ambiguous GT & 14 & 14 & java, & \includegraphics[height=0.6cm]{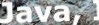} \\
    
    unclear GT & 5 & 20 & the &	\includegraphics[height=0.6cm, width=1.5cm]{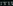}\\ 
    
    wrong weak label - punctuation & 1 & 2 &  1:15,000 &  \includegraphics[height=0.6cm, width=2.3cm]{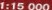}\\ 
    
    wrong weak label - other & 0 & 7 & hilling services &  \includegraphics[height=0.6cm, width=2.5cm]{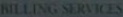}\\     
    
    \hline
    \end{tabular}
    \caption{PGT errors and ambiguities in the $\text{ABC}_3$ and $\text{UT}_6$ datasets.
    Counted on 500 sample crops from each.}
    \label{tab:pgt_error_freqs}
\end{table*}
}

\end{document}